\documentclass[sigconf]{acmart}

\AtBeginDocument{%
  \providecommand\BibTeX{{%
    \normalfont B\kern-0.5em{\scshape i\kern-0.25em b}\kern-0.8em\TeX}}}

\copyrightyear{2022}
\acmYear{2022}
\setcopyright{acmcopyright}\acmConference[MM '22]{Proceedings of the 30th ACM International Conference on Multimedia}{October 10--14, 2022}{Lisboa, Portugal}
\acmBooktitle{Proceedings of the 30th ACM International Conference on Multimedia (MM '22), October 10--14, 2022, Lisboa, Portugal}
\acmPrice{15.00}
\acmDOI{10.1145/3503161.3548059}
\acmISBN{978-1-4503-9203-7/22/10}

\usepackage{bbding} 
\usepackage{multirow} 
\usepackage{algorithm} 
\usepackage{bbm} 
\usepackage{algpseudocode}

\begin{document}

\title{Label-Efficient Domain Generalization via Collaborative Exploration and Generalization}


\author{Junkun Yuan}
\authornote{Both authors contributed equally to this research.}
\orcid{0000-0003-0012-7397}
\author{Xu Ma}
\authornotemark[1]
\affiliation{%
  \institution{Zhejiang University}
  \city{Hangzhou}
  \country{China}
}
\email{yuanjk@zju.edu.cn}
\email{maxu@zju.edu.cn}



\author{Defang Chen}
\affiliation{%
  \institution{Zhejiang University}
  \city{Hangzhou}
  \country{China}}
\email{defchern@zju.edu.cn}

\author{Kun Kuang}
\authornote{Corresponding author}
\affiliation{%
  \institution{Zhejiang University}
  \city{Hangzhou}
  \country{China}
}
\affiliation{%
  \institution{Shanghai AI Laboratory}
  \city{Shanghai}
  \country{China}
}
\email{kunkuang@zju.edu.cn}

\author{Fei Wu}
\affiliation{%
 \institution{Zhejiang University}
 \city{Hangzhou}
 \country{China}
}
\affiliation{%
 \institution{Shanghai Institute for Advanced Study of Zhejiang University}
 \city{Shanghai}
 \country{China}
}
\email{wufei@zju.edu.cn}

\author{Lanfen Lin}
\affiliation{%
  \institution{Zhejiang University}
  \city{Hangzhou}
  \country{China}
}
\email{llf@zju.edu.cn}


\renewcommand{\shortauthors}{Junkun Yuan et al.}



\begin{abstract}
Considerable progress has been made in domain generalization (DG) which aims to learn a generalizable model from multiple well-annotated source domains to unknown target domains. However, it can be prohibitively expensive to obtain sufficient annotation for source datasets in many real scenarios. To escape from the dilemma between domain generalization and annotation costs, in this paper, we introduce a novel task named label-efficient domain generalization (LEDG) to enable model generalization with label-limited source domains. To address this challenging task, we propose a novel framework called Collaborative Exploration and Generalization (CEG) which jointly optimizes active exploration and semi-supervised generalization. Specifically, in active exploration, to explore class and domain discriminability while avoiding information divergence and redundancy, we query the labels of the samples with the highest overall ranking of class uncertainty, domain representativeness, and information diversity. In semi-supervised generalization, we design MixUp-based intra- and inter-domain knowledge augmentation to expand domain knowledge and generalize domain invariance. We unify active exploration and semi-supervised generalization in a collaborative way and promote mutual enhancement between them, boosting model generalization with limited annotation. Extensive experiments show that CEG yields superior generalization performance. In particular, CEG can even use only 5\% data annotation budget to achieve competitive results compared to the previous DG methods with fully labeled data on PACS dataset.
\end{abstract}




\maketitle

\section{Introduction}
\label{sec:int}
Despite the remarkable success achieved by modern machine learning algorithms in visual recognition \cite{he2016deep, voulodimos2018deep, srinivas2021bottleneck}, it heavily relies on the i.i.d. assumption \cite{vapnik1992principles} that training and test datasets should have a consistent statistical pattern. Since machine learning systems are usually deployed in a wide range of scenarios where the test data are unknown in advance, it may inevitably result in serious model performance degradation when there exists a distinct \textit{distribution/domain shift} \cite{quionero2009dataset} between the training and test data.

\begin{figure}[t]
    \centering
    \includegraphics[trim={0cm 0cm 0cm 0cm},clip,width=0.99\columnwidth]{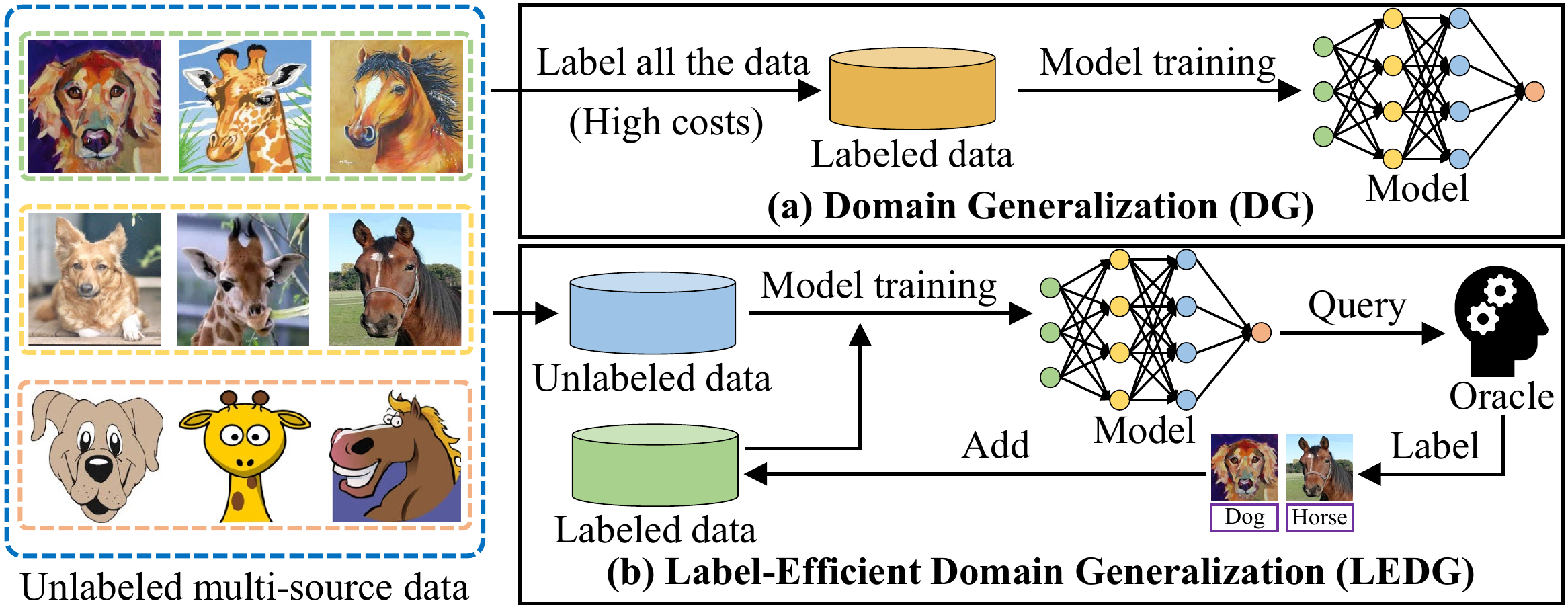}
    \caption{Comparison between the conventional DG (a) and the proposed LEDG (b) tasks. DG may need to label all source data. In comparison, LEDG queries the labels of a small quota of data with limited annotation budget and boosts domain generalization by exploiting both labeled and unlabeled data.}\label{fig:ledg}
\end{figure}

With awareness of this problem, \textit{domain generalization} (DG) \cite{blanchard2011generalizing} is introduced to extract domain invariance from multiple well-annotated source datasets/domains and train a generalizable model to unknown target domains. Lots of favorable DG algorithms \cite{shankar2018generalizing, Carlucci2019DomainGB, zhou2020deep, xu2021fourier, zhou2021domain, pandey2021generalization, dubey2021adaptive} have been proposed recently, however, these methods may need to be fed with a large amount of labeled multi-source data for identifying domain invariance and improving model generalization. It might impede the realization of the DG approaches in many real-world applications where labeling massive data could be expensive or even infeasible. 
For example, a highly accurate and robust system for the detection of lung lesion images of COVID-19 patients may demand a large number of labeled medical images from different hospitals as the source data for training \cite{ettinger2021large}, but it could be impractical to require numerous experienced clinicians to complete the annotation. 
Therefore, a \textit{dilemma} is encountered: The requirements of obtaining massive labeled source data for training a generalizable model may hard to be met in realistic scenarios due to the limited annotation budget. Meanwhile, without sufficient labeled data to provide adequate information of multi-source distribution, improving model generalization by identifying and learning the domain invariance is at serious risk of being misled. 

To escape from the dilemma, we introduce a more practical task named \textit{label-efficient domain generalization} (LEDG) to enable model generalization with label-limited source domains as shown in Figure \ref{fig:ledg}. 
Instead of getting fully labeled data, the LEDG task unleashes the potential of budget-limited annotation by querying the labels of a small quota of informative data, and leverages both the labeled and unlabeled data to improve domain generalization. 
LEDG permits the learning of generalizable models in real scenarios, but it could be much more challenging. 
The first challenge is from the clear domain distinctions that could exist in the multi-source data, which constitutes enormous obstacles for selecting the most informative samples and learning adequate information of multi-source distribution. Meanwhile, the second challenge is from the discrepant distributions that labeled and unlabeled data may be subject to, making the simultaneous utilization of them for extracting domain invariance and promoting model generalization extremely difficult. 

\textit{Active learning} (AL) \cite{wang2014new, sener2017active, ash2019deep, huang2021semi, kim2021task, joshi2009multi} and \textit{semi-supervised learning} (SSL) \cite{tarvainen2017mean, berthelot2019mixmatch, berthelot2020remixmatch, Sohn2020FixMatchSS} provide possible solutions to the introduced LEDG task. AL aims to query the labels of high-quality samples and SSL leverages the unlabeled data to improve performance with limited labeled data. However, the existing AL and SSL methods mostly depend on the i.i.d. assumption, hence may not be favorably extended to the generalization scenarios under distinct domain shifts. 
\textit{Semi-supervised domain generalization (SSDG)} \cite{zhou2021semi, wang2021better, yuan2021domain, liao2020deep, sharifi2020domain} tackles domain shift under the SSL setting. But some data directly assumed to be labeled in this task might not be helpful for generalization improvement but could increase the annotation costs.
Thus, it is imperative to find a solution to the challenging LEDG task for getting rid of the raised dilemma between domain generalization and annotation costs, realizing more practical training of generalizable models in real-world scenarios. 


To address the LEDG task, in this paper, we propose a novel framework called Collaborative Exploration and Generalization (CEG) which jointly optimizes active exploration and semi-supervised generalization. 
In active exploration, to unleash the power of the limited annotation, we query the labels of the samples with the highest overall ranking of class uncertainty, domain representativeness, and information diversity, exploring class and domain discriminability while avoiding information divergence and redundancy. 
In semi-supervised generalization, we augment intra- and inter-domain knowledge with MixUp \cite{zhang2017mixup} to expand domain knowledge and generalize domain invariance. An augmentation consistency constraint for unlabeled data and a prediction supervision for labeled data are further included to improve performance. 
We unify active exploration and semi-supervised generalization in a collaborative way by repeating them alternately, promoting closed-loop mutual enhancement between them for effective learning of domain invariance and label-efficient training of generalizable models. 

Our contributions are listed in the following.
(1) We introduce a more practical task named label-efficient domain generalization to permit generalization learning in real-world scenarios by tackling the dilemma between domain generalization and annotation costs.
(2) To solve this challenging task, we propose a semi-supervised active learning-based framework CEG to unify active query-based distribution exploration and semi-supervised training-based model generalization in a collaborative way, achieving closed-loop mutual enhancement between them.
(3) Extensive experiments show the superior generalization performance of CEG, which can even achieve competitive results using 5\% annotation budget compared to the previous DG methods with full annotation on PACS dataset.

\section{Related Work}
\label{sec:rel}

\textbf{Domain Generalization (DG).}
Different from domain adaptation (DA) \cite{wang2021interbn, deng2021informative, lv2021differentiated, yan2021pixel, huang2021few, ye2021source, li2021imbalanced, chen2021transferrable, ma2022attention, chen2022BA, chen2021multi, chen2021pareto} which adapts models from the source domain to the target, DG \cite{blanchard2011generalizing} assumes that the target domain is unknown during training and aims to train a generalizable model from the source domains. 
Increasing DG methods \cite{shankar2018generalizing, pandey2021generalization, dubey2021adaptive, volpi2021continual, mahajan2021domain, HuangWXH20, zhou2021domain3, yuan2021collaborative, yuan2021learning, kuang2018stable, kuang2022stable, kuang2021balance, kuang2020stable, shen2020stable} are proposed recently, they popularize various strategies via invariant representation learning \cite{zhao2020domain, dou2019domain, li2021progressive, li2018domain, qiao2020learning}, meta-learning \cite{shu2021open, balaji2018metareg, li2018learning, dou2019domain, li2019feature}, data augmentation \cite{Carlucci2019DomainGB, zhou2020deep, zhou2021domain, xu2021fourier, zhang2021deep, huang2021fsdr, jeon2021feature}, and others \cite{du2021learning, wang2021embracing, liu2021domain}. 
But they mostly need the fully labeled data to learn generalization.

\textbf{Semi-Supervised Domain Generalization (SSDG)} \cite{zhou2021semi, wang2021better, yuan2021domain, liao2020deep, sharifi2020domain} aims to reduce the reliance of DG on annotation via pseudo-labeling \cite{wang2021better}, consistency learning \cite{zhou2021semi}, or bias filtering \cite{yuan2021domain}. For example, StyleMatch \cite{zhou2021semi} combines consistency learning, model uncertainty learning, and style augmentation to utilize the annotation for improving model robustness. 
However, partial samples assumed to be labeled in the SSDG task may not be informative for boosting model generalization but increase the annotation costs. 

\textbf{Semi-Supervised Learning (SSL).}
SSL \cite{tarvainen2017mean, berthelot2019mixmatch, berthelot2020remixmatch, Sohn2020FixMatchSS, ZhangTYZKLZYW20} is a practicable way to use both labeled and unlabeled data. For example, MeanTeacher \cite{tarvainen2017mean} achieves significant performance by using the labeled data to optimize a student model, the prediction of which is constrained to be consistent with the prediction of a teacher model. 
But most of the SSL methods rely on the i.i.d. assumption, which can impair their generalization performance under domain shift. 

\begin{figure*}[t]
    \centering
    \includegraphics[trim={0cm 0cm 0cm 0cm},clip,width=1.99\columnwidth]{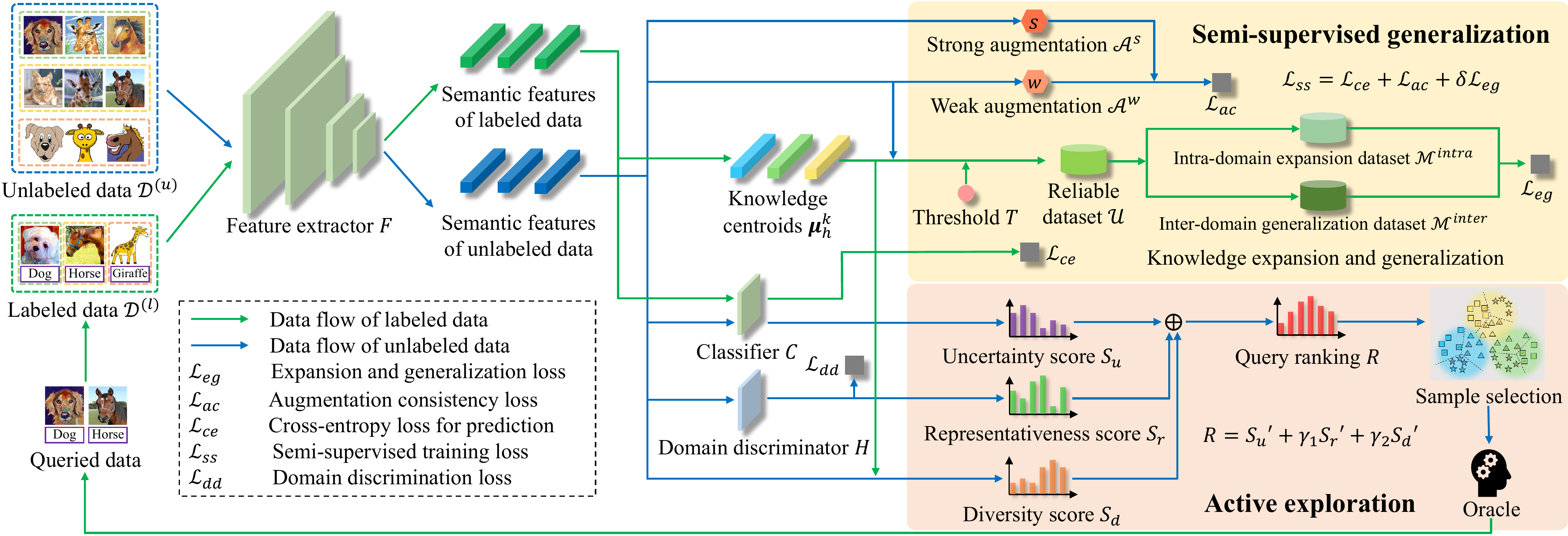}
    \caption{Overview of the Collaborative Exploration and Generalization (CEG) framework. In active exploration, samples with the highest overall ranking ($R$) of class uncertainty ($S_{u}$), domain representativeness ($S_{r}$), and information diversity ($S_{d}$) are selected to query for learning multi-source distribution. In semi-supervised generalization, expansion and generalization ($\mathcal{L}_{eg}$) of the known knowledge, and augmentation consistency ($\mathcal{L}_{ac}$) for unlabeled data and prediction supervision ($\mathcal{L}_{ce}$) for labeled data are devised to extract domain invariance and improve model generalization via semi-supervised training ($\mathcal{L}_{ss}$).}\label{fig:framework}
\end{figure*}

\textbf{Active Learning (AL).}
AL \cite{ash2019deep, wang2014new, joshi2009multi, sener2017active, kim2021task, huang2021semi} aims to select high-quality data for querying the labels. Pool-based AL \cite{ash2019deep, sener2017active, huang2021semi, kim2021task} is the most popular that chooses samples from an unlabeled pool and hand them over to the oracle to label. The labeled samples are then added to a labeled pool as newly acquired knowledge. 
Some successful uncertainty \cite{ash2019deep, wang2014new, joshi2009multi} and diversity \cite{ash2019deep, sener2017active} based methods select uncertain and diverse samples for learning task boundary and comprehensive information, respectively. However, the AL algorithms are mainly designed for single-domain data, thus may not be directly extended to the generalization scenarios. 



\section{Method}
\label{sec:met}

\subsection{Label-Efficient Domain Generalization}
We begin with the task setting of the introduced Label-Efficient Domain Generalization (LEDG). In the LEDG task, we have $K$ unlabeled source datasets $\{\mathcal{D}^{1},...,\mathcal{D}^{K}\}$ sampled from different data distributions $\{P(X^{1},Y^{1}),...,P(X^{K},Y^{K})\}$, respectively.
There are $N^{k}$ unlabeled data points being sampled for each dataset $\mathcal{D}^{k}$, i.e., $\mathcal{D}^{k}=\{\boldsymbol{x}_{i}^{k}\}_{i=1}^{N^{k}}$, for $k=1,...,K$. We further have an annotation budget $B$, i.e., the maximum number of samples that we are allowed to query their class labels. Each sample pair $(\boldsymbol{x},y)$ is defined on the image and label joint space $\mathcal{X}\times\mathcal{Y}$. Besides, the domain label $p_{i}^{k}$ for each sample $\boldsymbol{x}_{i}^{k}$ is given in our task. We consider a classification model $G$ composed of a feature extractor $F$ and a classifier head $C$, i.e., $G=C\circ{F}$. The goal of LEDG is to train the model $G$ by utilizing the unlabeled multi-source data $\{\mathcal{D}^{k}\}_{k=1}^{K}$ as well as the limited annotation budget $B$ for improving the generalization performance of the model on the target domains with unknown distributions. 
For convenience of the statement of our method, we denote the dataset consists of all the labeled (queried) samples $(\boldsymbol{x}_{i}^{(l)},y_{i}^{(l)})$ as $\mathcal{D}^{(l)}=\{(\boldsymbol{x}_{i}^{(l)},y_{i}^{(l)})\}_{i=1}^{N^{(l)}}$ and the dataset with all the unlabeled (not queried) samples $\boldsymbol{x}_{i}^{(u)}$ as $\mathcal{D}^{(u)}=\{\boldsymbol{x}_{i}^{(u)}\}_{i=1}^{N^{(u)}}$, where $N^{(l)}$ and $N^{(u)}$ are the data size of the labeled and unlabeled datasets, respectively. 
It is obvious that the whole data size $N^{(l)}+N^{(u)}=N^{1}+N^{2}+...+N^{K}$.


Our insight for this challenging task is to consider the labeled and unlabeled samples as the \textit{``known'' and ``unknown'' regions} of the multi-source distribution, respectively. 
In view of this, the core idea of our solution is to: (1) explore the key knowledge hidden in the unknown regions via active query for adequate multi-source distribution learning, (2) extract and generalize the domain invariance contained in the obtained knowledge in both the known and unknown regions via semi-supervised training, and (3) make the active query-based exploration and semi-supervised training-based generalization complement and promote each other to train a generalizable model. 
An overview of our framework, i.e., Collaborative Exploration and Generalization (CEG), is shown in Figure \ref{fig:framework}. 

\begin{figure*}[t]
    \centering
    \includegraphics[trim={0cm 0cm 0cm 0cm},clip,width=1.99\columnwidth]{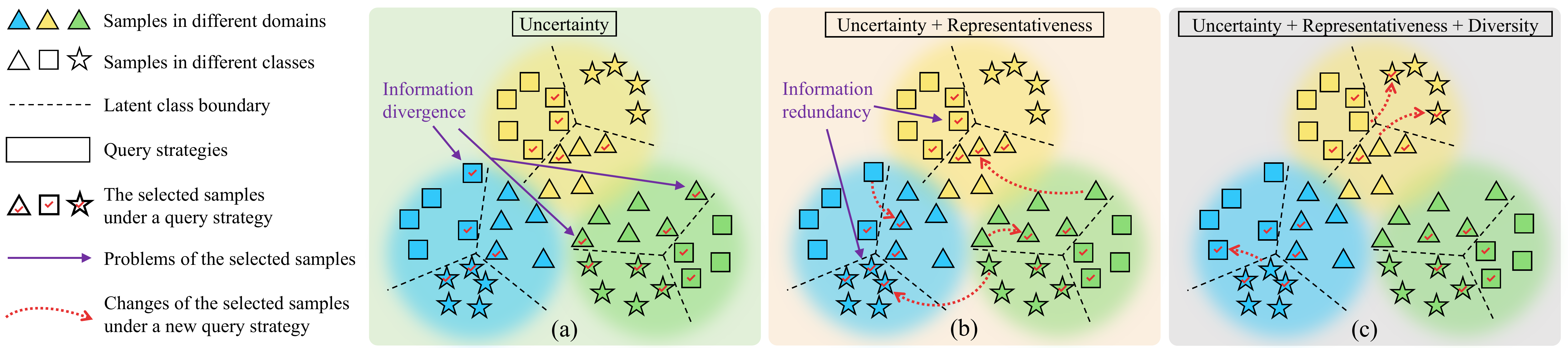}
    \caption{Comparisons of different query strategies for the active exploration. \textbf{(a):} Uncertainty criterion selects class-ambiguous samples but can cause information divergence, i.e., it may fail to select the representative samples for each domain. \textbf{(b):} The introduction of representativeness criterion helps to capture domain characteristics but can lead to information redundancy, i.e., it may select adjacent samples and hence reduce the efficiency of the limited annotation budget. \textbf{(c):} The further introduction of information diversity criterion guides the model to learn comprehensive knowledge of the data distribution.}\label{fig:active}
\end{figure*}

\subsection{Active Exploration}
There might be distinct domain divergence among the data distributions of the source domains. Meanwhile, each source domain contains the discriminative information of class boundary which is essential for the prediction task. Thus, we take class and domain discriminability as the key knowledge for learning the multi-source distribution in the active exploration. In light of this, we present to select the samples with high class uncertainty and domain representativeness. To avoid information redundancy, we further take information diversity into consideration. Figure \ref{fig:active} provides revealing insights into the elaborate strategy for the active exploration. 

To capture the key knowledge of class discriminability, we select the samples with high class uncertainty. Specifically, we adopt the margin of the top two model predictions to choose class-ambiguous samples to query. Let $G_{h}$ be the $h$-th dimension of the class prediction of the model $G$ (after $\mathrm{softmax}$ operation), then the class uncertainty score $S_{u}$ for each unlabeled sample $\boldsymbol{x}^{(u)}$ is defined as 
\begin{equation}\label{equ:unc}
    S_{u}(\boldsymbol{x}^{(u)})=\left(1-\left(\max_{h}{G_{h}\left(\boldsymbol{x}^{(u)}\right)}-\max_{h'|h'\neq{h}}{G_{h'}\left(\boldsymbol{x}^{(u)}\right)}\right)\right).
\end{equation}
We tend to query the samples with high uncertainty scores for their class-ambiguous. Labeling these samples provides the key knowledge of class discriminability, which helps the model to figure out class boundary and boosts its performance of class prediction. 

Different from the single-domain scenario considered in the AL methods \cite{wang2014new, joshi2009multi}, multiple source domains may lead to an information divergence problem here, i.e., the selected high-uncertainty samples are scattered at the domain boundary (see Figure \ref{fig:active} (a)). 
To sufficiently explore and grasp information of the multi-source distribution, the selected samples are required to represent the characteristics of each source domain. Therefore, given domain label $p^{(u)}$ of each unlabeled sample $\boldsymbol{x}^{(u)}$,
we first train a domain discriminator model $H$ with a domain discriminability loss:
\begin{equation}\label{equ:ldom}
    \mathcal{L}_{dd}=\mathbb{E}_{\boldsymbol{x}^{(u)}\in\mathcal{D}^{(u)}}\ell(H(\boldsymbol{x}^{(u)}),p^{(u)}),
\end{equation}
where $\ell$ is the cross-entropy loss. 
Let $H_{h}$ be the $h$-th dimension of the domain prediction of the model $H$, we then define domain representativeness score $S_{r}$ for each unlabeled sample $\boldsymbol{x}^{(u)}$ as:
\begin{equation}\label{equ:dom}
    S_{r}(\boldsymbol{x}^{(u)})=\max_{h}{H_{h}(\boldsymbol{x}^{(u)})}.
\end{equation}
Note that different from the class discriminability learning with class ambiguous data, here, we select the samples with high representativeness score, i.e., high domain confidence. It prevents the model from learning class discriminability in the remote areas of each source domain and hence loosing domain characteristics.

Then an information redundancy problem has arisen, i.e., the selected samples with high class uncertainty and domain representativeness may gather together (see Figure \ref{fig:active} (b)), which wastes the limited annotation budget. To disperse the information, we choose the samples that are far away from the known domain-class knowledge of the labeled data.
We make a knowledge dataset $\mathcal{K}_{h}^{k}$ be composed of the labeled data $(\boldsymbol{x}^{(l)},y^{(l)})\in\mathcal{D}^{(l)}$ that belongs to domain $k$ and class $h$ (if there is no such a sample in $\mathcal{D}^{(l)}$, then $\mathcal{K}_{h}^{k}=\emptyset$). 
Let $|\mathcal{K}_{h}^{k}|$ be the number of samples in the knowledge dataset $\mathcal{K}_{h}^{k}$ and $F$ be the feature extractor. We generate \textit{knowledge centroids} $\boldsymbol{\mu}_{h}^{k}$ for the known regions in the semantic feature space:
\begin{equation}\label{equ:anchor}
    \boldsymbol{\mu}_{h}^{k}=\frac{1}{|\mathcal{K}_{h}^{k}|}\sum_{(\boldsymbol{x}^{(l)},y^{(l)})\in\mathcal{K}_{h}^{k}}F(\boldsymbol{x}^{(l)}).
\end{equation}
We let a set $\mathcal{B}$ be composed of all the knowledge centroids $\boldsymbol{\mu}_{h}^{k}$ if $\mathcal{K}_{h}^{k}\neq{\emptyset}$. Then, we define information diversity score $S_{d}$ as
\begin{equation}\label{equ:div}
    S_{d}(\boldsymbol{x}^{(u)})=\min_{\boldsymbol{\mu}_{h}^{k}\in\mathcal{B}}\mathrm{dist}(\boldsymbol{x}^{(u)},\boldsymbol{\mu}_{h}^{k}),
\end{equation}
where $\mathrm{dist}(\cdot,\cdot)$ is a distance metric used as cosine distance in experiments. 
We tend to choose samples with high diversity score, i.e., far away from the closest centroids, facilitating the unknown exploration for comprehensive learning of multi-source distribution. 

To avoid numerical issues, we integrate uncertainty score $S_{u}$, representativeness score $S_{r}$, and diversity score $S_{d}$ by adopting their rankings, which we denote as $S_{u}'$, $S_{r}'$, and $S_{d}'$, respectively. Finally, we have an overall query ranking $R$ for each unlabeled sample $\boldsymbol{x}^{(u)}$:
\begin{equation}\label{equ:q}
    \begin{aligned}
        R(\boldsymbol{x}^{(u)})=&S_{u}'(\boldsymbol{x}^{(u)})+\gamma_{1}S_{r}'(\boldsymbol{x}^{(u)})+\gamma_{2}S_{d}'(\boldsymbol{x}^{(u)}),
    \end{aligned}
\end{equation}
where $\gamma_{1}$ and $\gamma_{2}$ are trade-off hyper-parameters. 
Note that we rank each score, i.e., $S_{u}$, $S_{r}$, and $S_{d}$, from high to low, for selecting the most informative samples of the multi-source data distribution.

\subsection{Semi-Supervised Generalization}
\begin{figure}[t]
    \centering
    \includegraphics[trim={0cm 0cm 0cm 0cm},clip,width=0.99\columnwidth]{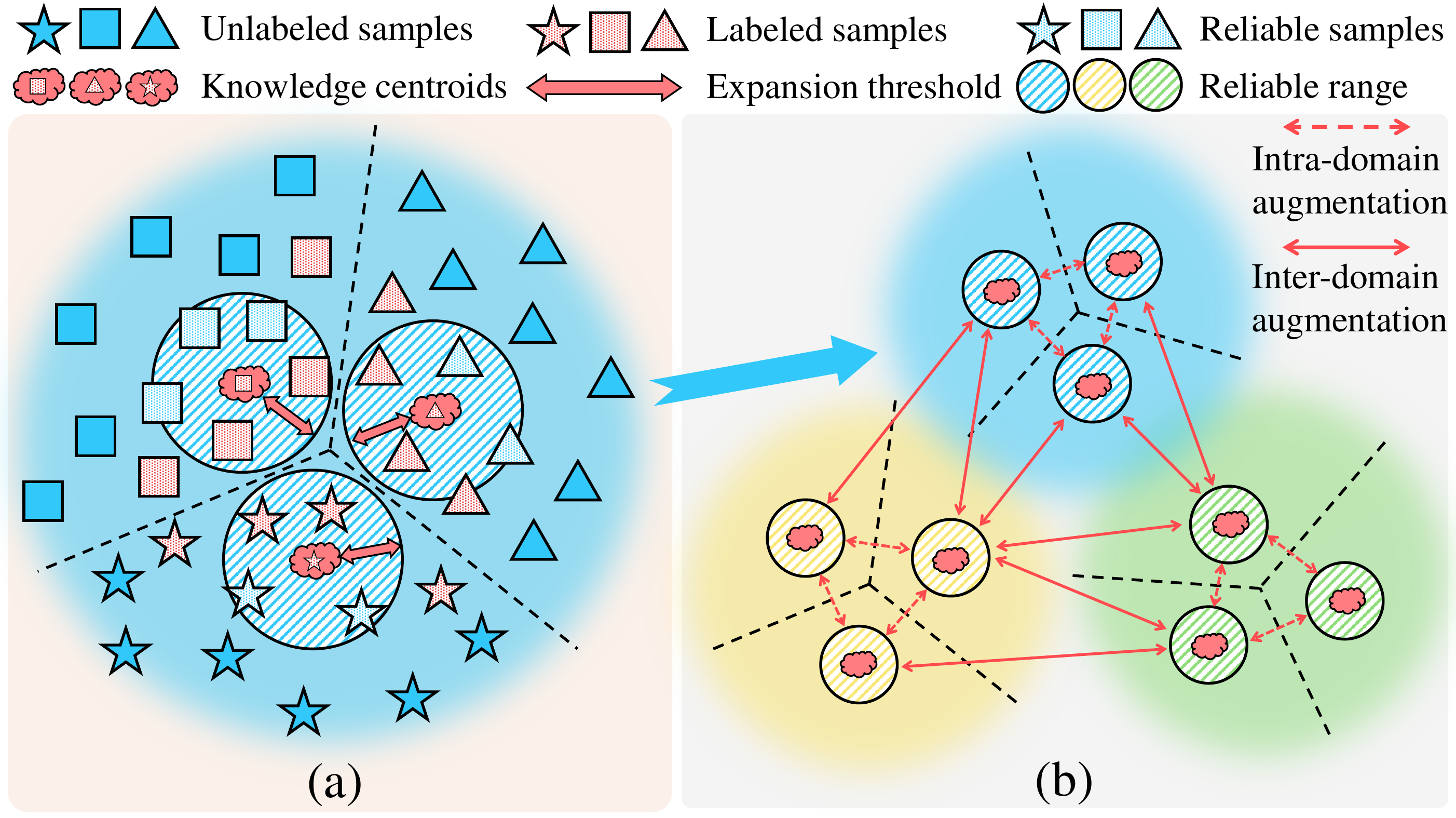}
    \caption{Semi-supervised generalization with knowledge augmentation. \textbf{(a)}: Unlabeled samples within reliable range (scaled by expansion threshold), i.e., the reliable samples, are pseudo-labeled by the nearest knowledge centroids, i.e., the centers of the labeled data. \textbf{(b)}: The reliable samples are then augmented in the same domain and across domains to expand knowledge and generalize domain invariance.}\label{fig:semi}
\end{figure}

With active query-based exploration, we have a small quota of labeled data and massive unlabeled data, i.e., small range of known regions and large range of unknown regions of the data distribution. In semi-supervised generalization, we aim to expand domain knowledge and learn domain invariance via MixUp-based intra- and inter-domain knowledge augmentation as shown in Figure \ref{fig:semi}. 

We start by defining the unlabeled samples that close to the knowledge centroids $\boldsymbol{\mu}_{h}^{k}$ as ``reliable samples'', and construct a reliable dataset $\mathcal{U}$ with expansion threshold $T$ to tune reliable range:
\begin{equation}\label{equ:I}
    \mathcal{U}=\{(\boldsymbol{x}^{(u)},y^{(u)})|\min_{\boldsymbol{\mu}_{h}^{k}\in\mathcal{B}}\mathrm{dist}(\boldsymbol{x}^{(u)},\boldsymbol{\mu}_{h}^{k})<T\},
\end{equation}
where pseudo label $y^{(u)}$ of each unlabeled sample $\boldsymbol{x}^{(u)}$ is assigned by the nearest knowledge centroids $\boldsymbol{\mu}_{h}^{k}$, that is,
\begin{equation}\label{equ:yu}
    y^{(u)}={\arg\min}_{h:\boldsymbol{\mu}_{h}^{k}\in\mathcal{B}}\mathrm{dist}(\boldsymbol{x}^{(u)},\boldsymbol{\mu}_{h}^{k}).
\end{equation}
A low threshold value $T$ leads to few reliable samples but high dependability of its pseudo labels, and vice versa. 
To arrange learning tasks in the order of difficulty for helping the model gain sufficient basic and easy knowledge before handling more complex data, we let $T$ increase with epochs and dynamically tune learning difficulty:
\begin{equation}\label{equ:T}
    T=T^{ini}+\frac{T^{fin}-T^{ini}}{E_{tot}}*E_{cur},
\end{equation}
where $T^{ini}$ and $T^{fin}$ are the initial and final threshold values, $E_{tot}$ and $E_{cur}$ are the total and current epochs, respectively. It makes the model expand knowledge stably with the high dependable samples at the beginning, and break through the hard samples gradually. 

We expand domain-class knowledge in each domain and across domains with the reliable dataset $\mathcal{U}$, and construct MixUp-based intra- and inter-domain knowledge augmentation datasets, i.e., $\mathcal{M}^{intra}$ and $\mathcal{M}^{inter}$, respectively. That is,
\begin{equation}\label{equ:intra}
    \begin{aligned}
        \mathcal{M}^{intra}=&\{(\lambda\boldsymbol{x}^{(u)}_{i}+(1-\lambda)\boldsymbol{x}^{(u)}_{j},\lambda y^{(u)}_{i}+(1-\lambda)y^{(u)}_{j})|\\
        &(\boldsymbol{x}_{i}^{(u)},y_{i}^{(u)})\in\mathcal{D}^{m},(\boldsymbol{x}_{j}^{(u)},y_{j}^{(u)})\in\mathcal{D}^{n},m=n\},
    \end{aligned}
\end{equation}
\begin{equation}\label{equ:inter}
    \begin{aligned}
        \mathcal{M}^{inter}=&\{(\lambda\boldsymbol{x}^{(u)}_{i}+(1-\lambda)\boldsymbol{x}^{(u)}_{j},\lambda y^{(u)}_{i}+(1-\lambda)y^{(u)}_{j})|\\
        &(\boldsymbol{x}_{i}^{(u)},y_{i}^{(u)})\in\mathcal{D}^{m},(\boldsymbol{x}_{j}^{(u)},y_{j}^{(u)})\in\mathcal{D}^{n},m\neq n\},
    \end{aligned}
\end{equation}
where $\lambda\sim Beta(\alpha,\alpha)$ with $\alpha=0.2$ as in \cite{zhang2017mixup}. $\mathcal{M}^{intra}$ and $\mathcal{M}^{inter}$ open up the association among known regions within and across domains, respectively. 
To broaden the known regions and learn domain-invariant representations for improving out-of-domain generalization ability, we train the model on the union of the augmented datasets by optimizing an expansion and generalization loss $\mathcal{L}_{eg}$:
\begin{equation}\label{equ:ep}
    \mathcal{L}_{eg}=\mathbb{E}_{(\boldsymbol{x}^{(u)},y^{(u)})\in\mathcal{M}^{intra}\cup\mathcal{M}^{inter}}\ell(G(\boldsymbol{x}^{(u)}),y^{(u)}).
\end{equation}

\begin{algorithm}[t]
    \caption{Collaborative Exploration and Generalization}  
    \label{alg: method}
    \begin{algorithmic}[1]
    \Require 
    Source datasets $\{\mathcal{D}^{k}\}_{k=1}^{K}$;
    Pretraining epochs $N_{p}$ and learning epochs $N_{l}$; Annotation budget $B$ and initial budget $B^{ini}$.
    \Ensure A well trained prediction model $\hat{G}$; 
    \State Initialize labeled dataset $\mathcal{D}^{(l)}$ with $B^{ini}$;
    \For{$n=1$ to $N_{l}$}
        \State Get knowledge centroids $\boldsymbol{\mu}_{h}^{k}$ with $\mathcal{D}^{(l)}$ via Equation (\ref{equ:anchor});
        \If{$n>N_{p}$}  // begin to query after pretraining
            \State Select a subset $\Delta\mathcal{D}^{(u)}$ from $\mathcal{D}^{(u)}$ via Equation (\ref{equ:q});
            \State Query labels of $\Delta\mathcal{D}^{(u)}$ for  $\Delta\mathcal{D}^{(u)}\rightarrow\Delta\mathcal{D}^{(l)}$;
            \State $\mathcal{D}^{(l)}\leftarrow\mathcal{D}^{(l)}\cup\Delta\mathcal{D}^{(l)}$, $\mathcal{D}^{(u)}\leftarrow\mathcal{D}^{(u)}/\Delta\mathcal{D}^{(u)}$;
        \EndIf
        \State Train the discriminator $H$ with $\mathcal{D}^{(u)}$ via Equation (\ref{equ:ldom});
        \State Train the model $G$ with $\mathcal{D}^{(l)}$ and $\mathcal{D}^{(u)}$ via Equation (\ref{equ:L}).
    \EndFor
    \end{algorithmic}
\end{algorithm}

We further utilize the unknown and known regions by adopting augmentation consistency constraint \cite{Sohn2020FixMatchSS} for the unlabeled data and prediction supervision for the labeled data, respectively. 
Let $\mathcal{A}^{w}$ and $\mathcal{A}^{s}$ be weak (flip-and-shift) and strong \cite{cubuk2019autoaugment} augmentation functions, respectively. Pseudo labels $\hat{q}$ can be assigned by $\mathcal{A}^{w}$ via $\hat{q}={\arg\max}_{h}G_{h}(\mathcal{A}^{w}(\boldsymbol{x}^{(u)}))$. The augmentation consistency loss $\mathcal{L}_{ac}$ makes the model predictions of strong augmented data, i.e., $G(\mathcal{A}^{s}(\boldsymbol{x}^{(u)}))$, and weak augmented label, i.e., $\hat{q}$, to be consistent:
\begin{equation}
    \mathcal{L}_{ac}=\mathbb{E}_{\boldsymbol{x}^{(u)}\in\mathcal{D}^{(u)}}I(\boldsymbol{x}^{(u)})\ell(G(\mathcal{A}^{s}(\boldsymbol{x}^{(u)})), \hat{q}),
\end{equation}
where an indicator $I(\boldsymbol{x}^{(u)})=\mathbbm{1}(\max_{h}{(G_{h}(\mathcal{A}^{w}(\boldsymbol{x}^{(u)}))}\geq\tau)$ ($\tau$ is set to 0.95 as in \cite{Sohn2020FixMatchSS}) selects high dependable data. This constraint helps the model to capture structural knowledge in the unknown regions via unsupervised learning. 
For prediction supervision, we adopt a cross-entropy classification loss $\mathcal{L}_{ce}$ for the labeled data: 
\begin{equation}\label{equ:ce}
    \mathcal{L}_{ce}=\mathbb{E}_{(\boldsymbol{x}^{(l)},y^{(l)})\in\mathcal{D}^{(l)}}\ell(G(\boldsymbol{x}^{(l)}),y^{(l)}).
\end{equation}
A semi-supervised training loss $\mathcal{L}_{ss}$ is then derived as:
\begin{equation}\label{equ:L}
    \mathcal{L}_{ss}=\mathcal{L}_{ce}+\mathcal{L}_{ac}+\delta\mathcal{L}_{eg},
\end{equation}
where $\delta$ is a hyper-parameter of knowledge expansion and generalization. We set the weights of $\mathcal{L}_{ce}$ and $\mathcal{L}_{ac}$ to 1 as in \cite{Sohn2020FixMatchSS}. 

Our framework CEG explores informative unlabeled samples for learning key knowledge of multi-source distribution using limited annotation budget, promoting the expansion and generalization of the key knowledge in semi-supervised training. Then a well trained model is continuously utilized to select more effective samples in the next round of query. The active exploration and semi-supervised generalization are unified in a collaborative way by being repeated alternately. They complement and promote each other to enable label-efficient domain generalization. The learning process of CEG is stated in Algorithm \ref{alg: method}. Note that we use an initial budget $B^{ini}$ from the annotation budget $B$ to initialize the labeled dataset $D^{(l)}$ via uniform sample selection, and pretrain the model before active query to solve a cold start problem, to our empirical experience.

\section{Experiments}
\label{sec:exp}

\begin{table*}[t]
\caption{Performance (\%) comparisons between \textbf{CEG} and \textbf{DG methods} on PACS dataset with 5\% annotation budget. The results with fully labeled source data (100\% annotation) are given in parentheses. The best results are emphasized in bold.}
\label{table:dg-pacs}
\centering
\resizebox{1\linewidth}{!}{
\begin{tabular}{l|cccc|c}
\toprule
Methods & Art & Cartoon & Photo & Sketch & Average \\
\midrule
DeepAll & 55.79{\scriptsize{$\pm$2.92}} (73.59{\scriptsize{$\pm$2.89}}) & 61.84{\scriptsize{$\pm$1.98}} (70.63{\scriptsize{$\pm$2.33}}) & 80.21{\scriptsize{$\pm$1.14}} (89.36{\scriptsize{$\pm$1.21}}) & 63.00{\scriptsize{$\pm$1.60}} (80.06{\scriptsize{$\pm$0.95}}) & 65.21{\scriptsize{$\pm$0.46}} (78.41{\scriptsize{$\pm$0.44}}) \\
JiGen \cite{Carlucci2019DomainGB} &  53.03{\scriptsize{$\pm$5.19}} (77.05{\scriptsize{$\pm$1.83}}) & 55.09{\scriptsize{$\pm$1.85}} (76.16{\scriptsize{$\pm$2.02}}) & 78.62{\scriptsize{$\pm$5.83}} (93.81{\scriptsize{$\pm$1.51}}) & {22.28\scriptsize{$\pm$4.05}} (70.93{\scriptsize{$\pm$0.72}}) & 52.26{\scriptsize{$\pm$0.25}} (79.49{\scriptsize{$\pm$0.67}}) \\
FACT \cite{xu2021fourier} &  74.21{\scriptsize{$\pm$0.30}} (84.76{\scriptsize{$\pm$0.77}}) & 65.82{\scriptsize{$\pm$2.03}} (77.52{\scriptsize{$\pm$0.70}}) & 90.48{\scriptsize{$\pm$1.17}} (95.29{\scriptsize{$\pm$0.31}}) & 55.24{\scriptsize{$\pm$3.62}} (78.97{\scriptsize{$\pm$0.32}}) & 71.44{\scriptsize{$\pm$1.02}} (84.13{\scriptsize{$\pm$0.24}}) \\ 
DDAIG \cite{zhou2020deep} & 62.87{\scriptsize{$\pm$3.13}} (77.80{\scriptsize{$\pm$1.09}}) & 57.64{\scriptsize{$\pm$6.59}} (75.35{\scriptsize{$\pm$3.21}}) & 81.48{\scriptsize{$\pm$3.82}} (89.66{\scriptsize{$\pm$1.74}}) & 36.61{\scriptsize{$\pm$2.23}} (73.70{\scriptsize{$\pm$2.99}}) & 59.65{\scriptsize{$\pm$2.22}} (79.13{\scriptsize{$\pm$0.91}}) \\
RSC \cite{HuangWXH20} & 59.57{\scriptsize{$\pm$3.37}} (77.88{\scriptsize{$\pm$0.66}}) & 59.61{\scriptsize{$\pm$3.22}} (73.90{\scriptsize{$\pm$2.12}}) & 84.43{\scriptsize{$\pm$3.59}} (93.85{\scriptsize{$\pm$0.80}}) & 57.38{\scriptsize{$\pm$6.61}} (80.66{\scriptsize{$\pm$0.81}}) & 65.25{\scriptsize{$\pm$2.95}} (81.57{\scriptsize{$\pm$0.71}}) \\
CrossGrad \cite{shankar2018generalizing} & 56.06{\scriptsize{$\pm$7.04}} (75.69{\scriptsize{$\pm$2.25}}) & 52.73{\scriptsize{$\pm$4.17}} (76.51{\scriptsize{$\pm$3.24}}) & 80.51{\scriptsize{$\pm$1.97}} (91.33{\scriptsize{$\pm$0.50}}) & 41.25{\scriptsize{$\pm$5.21}} (70.50{\scriptsize{$\pm$0.97}}) & 57.64{\scriptsize{$\pm$2.13}} (78.51{\scriptsize{$\pm$1.40}}) \\
DAEL \cite{zhou2021domain3} & 66.24{\scriptsize{$\pm$1.86}} (83.51{\scriptsize{$\pm$0.83}}) & 61.72{\scriptsize{$\pm$1.89}} (72.31{\scriptsize{$\pm$2.67}}) & 89.98{\scriptsize{$\pm$0.37}} (95.74{\scriptsize{$\pm$0.08}}) & 32.50{\scriptsize{$\pm$2.20}} (78.87{\scriptsize{$\pm$0.59}}) & 62.61{\scriptsize{$\pm$0.46}} (82.61{\scriptsize{$\pm$0.98}}) \\
\midrule
\textbf{CEG (ours)} & \textbf{80.12}\scriptsize{$\pm$0.37} & \textbf{71.11}\scriptsize{$\pm$0.96} & \textbf{92.32}\scriptsize{$\pm$1.68} & \textbf{73.13}\scriptsize{$\pm$2.87} & \textbf{79.17}\scriptsize{$\pm$0.83} \\
\bottomrule
\end{tabular}}
\end{table*}

\begin{table*}[t]
\caption{Performance (\%) comparisons between \textbf{CEG} and \textbf{DG methods} on Office-Home dataset with 5\% annotation budget. The results with fully labeled source data (100\% annotation) are given in parentheses. The best results are emphasized in bold.}
\label{table:dg-home}
\centering
\resizebox{1\linewidth}{!}{
\begin{tabular}{l|cccc|c}
\toprule
Methods & Art & Clipart & Product & Real-World & Average \\
\midrule
DeepAll & 34.73{\scriptsize{$\pm$1.13}} (47.06{\scriptsize{$\pm$1.35}}) & 34.46{\scriptsize{$\pm$2.11}} (47.50{\scriptsize{$\pm$0.91}}) & 46.20{\scriptsize{$\pm$1.51}} (64.89{\scriptsize{$\pm$0.65}}) & 48.89{\scriptsize{$\pm$0.87}} (65.16{\scriptsize{$\pm$0.62}}) & 41.07{\scriptsize{$\pm$0.93}} (56.15{\scriptsize{$\pm$0.59}}) \\
JiGen \cite{Carlucci2019DomainGB} & 29.62{\scriptsize{$\pm$2.25}} (52.67{\scriptsize{$\pm$0.95}}) & 25.52{\scriptsize{$\pm$2.12}} (50.40{\scriptsize{$\pm$0.97}}) & 37.91{\scriptsize{$\pm$1.33}} (71.21{\scriptsize{$\pm$0.12}}) & 39.84{\scriptsize{$\pm$0.61}} (72.24{\scriptsize{$\pm$0.15}}) & 33.22{\scriptsize{$\pm$0.93}} (61.63{\scriptsize{$\pm$0.25}}) \\
FACT \cite{xu2021fourier} & 40.71{\scriptsize{$\pm$0.08}} (58.98{\scriptsize{$\pm$0.29}}) & 32.12{\scriptsize{$\pm$0.17}} (53.53{\scriptsize{$\pm$0.35}}) & 48.05{\scriptsize{$\pm$0.14}} (74.47{\scriptsize{$\pm$0.56}}) & 49.16{\scriptsize{$\pm$0.17}} (75.63{\scriptsize{$\pm$0.67}}) & 42.51{\scriptsize{$\pm$0.09}} (65.65{\scriptsize{$\pm$0.41}}) \\ 
DDAIG \cite{zhou2020deep} & 35.20{\scriptsize{$\pm$1.06}} (55.05{\scriptsize{$\pm$0.69}}) & 29.75{\scriptsize{$\pm$0.50}} (52.37{\scriptsize{$\pm$0.58}}) & 42.42{\scriptsize{$\pm$0.58}} (72.00{\scriptsize{$\pm$0.58}}) & 43.07{\scriptsize{$\pm$0.12}} (73.54{\scriptsize{$\pm$0.19}}) & 37.61{\scriptsize{$\pm$0.16}} (63.24{\scriptsize{$\pm$0.35}}) \\
RSC \cite{HuangWXH20} & 31.95{\scriptsize{$\pm$1.24}} (56.06{\scriptsize{$\pm$0.71}}) & 28.62{\scriptsize{$\pm$1.53}} (52.95{\scriptsize{$\pm$0.31}}) & 40.88{\scriptsize{$\pm$1.87}} (72.61{\scriptsize{$\pm$0.39}}) & 42.43{\scriptsize{$\pm$0.69}} (73.42{\scriptsize{$\pm$0.38}}) & 35.97{\scriptsize{$\pm$0.61}} (63.76{\scriptsize{$\pm$0.25}}) \\
CrossGrad \cite{shankar2018generalizing} & 35.05{\scriptsize{$\pm$0.37}} (54.42{\scriptsize{$\pm$0.55}}) & 30.86{\scriptsize{$\pm$1.74}} (52.63{\scriptsize{$\pm$0.77}}) & 45.10{\scriptsize{$\pm$1.76}} (73.00{\scriptsize{$\pm$0.47}}) & 44.41{\scriptsize{$\pm$2.08}} (73.42{\scriptsize{$\pm$0.74}}) & 38.86{\scriptsize{$\pm$0.27}} (63.37{\scriptsize{$\pm$0.24}}) \\
DAEL \cite{zhou2021domain3} & 35.93{\scriptsize{$\pm$0.57}} (59.20{\scriptsize{$\pm$0.56}}) & 30.71{\scriptsize{$\pm$0.86}} (50.97{\scriptsize{$\pm$2.63}}) & 42.79{\scriptsize{$\pm$0.99}} (73.53{\scriptsize{$\pm$0.52}}) & 43.95{\scriptsize{$\pm$0.86}} (76.56{\scriptsize{$\pm$0.45}}) & 38.35{\scriptsize{$\pm$0.12}} (65.06{\scriptsize{$\pm$0.55}}) \\
\midrule
\textbf{CEG (ours)} & \textbf{47.60}\scriptsize{$\pm$1.32} & \textbf{42.01}\scriptsize{$\pm$1.19} & \textbf{56.20}\scriptsize{$\pm$1.79} & \textbf{57.69}\scriptsize{$\pm$1.18} & \textbf{50.87}\scriptsize{$\pm$0.99}\\
\bottomrule
\end{tabular}}
\end{table*}

\begin{table*}[t]
\caption{Comparisons between \textbf{CEG} and \textbf{AL, SSL, SSDG methods} on PACS and Office-Home datasets with \textbf{5\%} annotation budget.}
\label{table:result}
\centering
\resizebox{1\linewidth}{!}{
\begin{tabular}{l|cccc|c||cccc|c}
\toprule
\multirow{2}{*}{Methods} & \multicolumn{5}{c||}{PACS dataset (\%)} & \multicolumn{5}{c}{Office-Home dataset (\%)} \\
\cline{2-11}
& Art & Cartoon & Photo & Sketch & Average & Art & Clipart & Product & Real-World & Average \\
\midrule
Uniform  & 55.56\scriptsize{$\pm$2.92} & 61.61\scriptsize{$\pm$1.98} & 79.98\scriptsize{$\pm$1.14} & 62.77\scriptsize{$\pm$1.60} & 64.98\scriptsize{$\pm$0.46} 
& 34.27\scriptsize{$\pm$1.13} & 34.00\scriptsize{$\pm$2.11} & 45.74\scriptsize{$\pm$1.51} & 48.43\scriptsize{$\pm$0.87} & 40.61\scriptsize{$\pm$0.93} \\
Entropy \cite{wang2014new} & 58.79\scriptsize{$\pm$2.36} & 63.49\scriptsize{$\pm$1.78} & 82.47\scriptsize{$\pm$0.99} & 61.67\scriptsize{$\pm$0.82} & 66.61\scriptsize{$\pm$0.58} & 
34.06\scriptsize{$\pm$1.59} & 32.02\scriptsize{$\pm$1.84} & 46.72\scriptsize{$\pm$0.99} & 47.03\scriptsize{$\pm$0.96} & 39.96\scriptsize{$\pm$0.90}\\
BvSB \cite{joshi2009multi} & 62.85\scriptsize{$\pm$1.83} & 63.17\scriptsize{$\pm$1.02} & 79.57\scriptsize{$\pm$3.46} & 63.61\scriptsize{$\pm$5.97} & 67.30\scriptsize{$\pm$0.97} & 
35.58\scriptsize{$\pm$1.44} & 35.32\scriptsize{$\pm$2.29} & 47.66\scriptsize{$\pm$1.42} & 50.39\scriptsize{$\pm$0.54} & 42.24\scriptsize{$\pm$0.82}\\
Confidence \cite{wang2014new} & 58.02\scriptsize{$\pm$2.14} & 59.48\scriptsize{$\pm$1.61} & 81.75\scriptsize{$\pm$3.40} & 61.04\scriptsize{$\pm$2.86} & 65.07\scriptsize{$\pm$1.60} 
& 36.35\scriptsize{$\pm$1.23} & 36.21\scriptsize{$\pm$1.24} & 47.88\scriptsize{$\pm$0.75} & 50.46\scriptsize{$\pm$2.20} & 42.73\scriptsize{$\pm$0.83}\\
CoreSet \cite{sener2017active} & 61.48\scriptsize{$\pm$4.57} & 58.74\scriptsize{$\pm$2.66} & 79.03\scriptsize{$\pm$3.71} & 60.61\scriptsize{$\pm$2.25} & 64.96\scriptsize{$\pm$1.06} 
& 37.54\scriptsize{$\pm$0.76} & 35.75\scriptsize{$\pm$2.55} & 49.44\scriptsize{$\pm$1.21} & 51.06\scriptsize{$\pm$1.75} & 43.45\scriptsize{$\pm$0.66}\\
BADGE \cite{ash2019deep} & 54.49\scriptsize{$\pm$1.67} & 63.10\scriptsize{$\pm$1.60} & 80.84\scriptsize{$\pm$1.19} & 65.57\scriptsize{$\pm$5.99} & 66.50\scriptsize{$\pm$2.11}
& 37.81\scriptsize{$\pm$1.07} & 36.86\scriptsize{$\pm$2.62} & 49.90\scriptsize{$\pm$2.00} & 51.26\scriptsize{$\pm$2.77} & 43.96\scriptsize{$\pm$0.82}\\
MeanTeacher \cite{tarvainen2017mean} & 53.84\scriptsize{$\pm$6.41} & 54.86\scriptsize{$\pm$4.14} & 78.86\scriptsize{$\pm$4.63} & 35.52\scriptsize{$\pm$4.57} & 55.77\scriptsize{$\pm$1.43} 
& 32.70\scriptsize{$\pm$1.56} & 27.25\scriptsize{$\pm$2.92} & 43.01\scriptsize{$\pm$2.04} & 42.41\scriptsize{$\pm$4.03} & 36.35\scriptsize{$\pm$1.25}\\
MixMatch \cite{berthelot2019mixmatch} & 63.92\scriptsize{$\pm$1.77} & 61.37\scriptsize{$\pm$2.31} & 81.14\scriptsize{$\pm$4.12} & 55.46\scriptsize{$\pm$0.61} & 65.47\scriptsize{$\pm$1.70} 
& 25.65\scriptsize{$\pm$0.66} & 22.90\scriptsize{$\pm$2.24} & 33.80\scriptsize{$\pm$0.93} & 28.35\scriptsize{$\pm$2.24} & 27.68\scriptsize{$\pm$0.80}\\
FixMatch \cite{Sohn2020FixMatchSS} & 78.60\scriptsize{$\pm$1.47} & 71.14\scriptsize{$\pm$2.49} & 92.17\scriptsize{$\pm$1.02} & 69.16\scriptsize{$\pm$0.94} & 77.77\scriptsize{$\pm$0.91} 
& 36.76\scriptsize{$\pm$1.84} & 31.09\scriptsize{$\pm$2.53} & 44.79\scriptsize{$\pm$4.20} & 45.07\scriptsize{$\pm$5.18} & 39.43\scriptsize{$\pm$2.21}\\
StyleMatch \cite{zhou2021semi}  & 72.67\scriptsize{$\pm$1.08} & \textbf{73.07}\scriptsize{$\pm$0.81} & 89.61\scriptsize{$\pm$0.74} & \textbf{76.46}\scriptsize{$\pm$0.93} & 77.95\scriptsize{$\pm$0.56} 
& 42.01\scriptsize{$\pm$0.68} & 40.95\scriptsize{$\pm$0.97} & 47.65\scriptsize{$\pm$1.70} & 51.93\scriptsize{$\pm$0.26} & 45.63\scriptsize{$\pm$0.29}\\
\midrule
\textbf{CEG (ours)} & \textbf{80.12}\scriptsize{$\pm$0.37} & 71.11\scriptsize{$\pm$0.96} & \textbf{92.32}\scriptsize{$\pm$1.68} & 73.13\scriptsize{$\pm$2.87} & \textbf{79.17}\scriptsize{$\pm$0.83} & \textbf{47.60}\scriptsize{$\pm$1.32} & \textbf{42.01}\scriptsize{$\pm$1.19} & \textbf{56.20}\scriptsize{$\pm$1.79} & \textbf{57.69}\scriptsize{$\pm$1.18} & \textbf{50.87}\scriptsize{$\pm$0.99}\\
\bottomrule
\end{tabular}}
\end{table*}

\begin{figure*}[t]
    \centering
    \includegraphics[trim={0cm 0cm 0cm 0cm},clip,width=1.99\columnwidth]{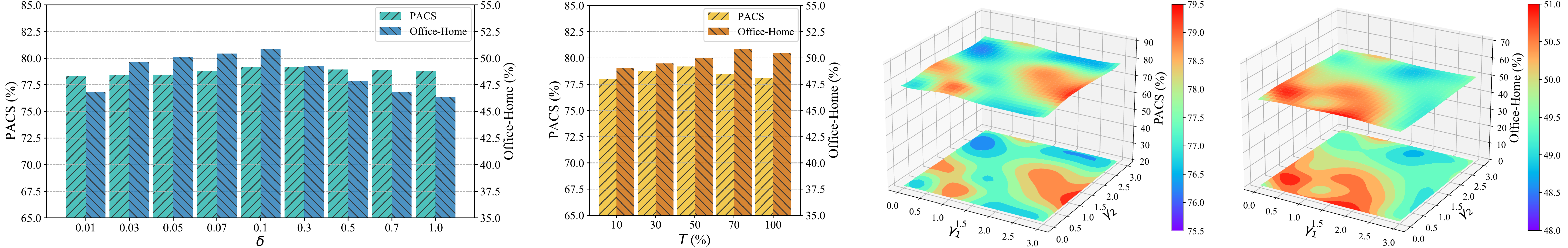}
    \caption{\textbf{Sensitivity analysis of hyper-parameters.} From left to right: $\delta$, $T$, $\gamma_{1}$ and $\gamma_{2}$ on PACS, $\gamma_{1}$ and $\gamma_{2}$ on Office-Home.}\label{fig:sen}
\end{figure*}

\def\curvesize{0.4}
\begin{figure}[t]
    \centering
    \includegraphics[trim={0cm 0cm 0cm 0cm},clip,width=0.99\columnwidth]{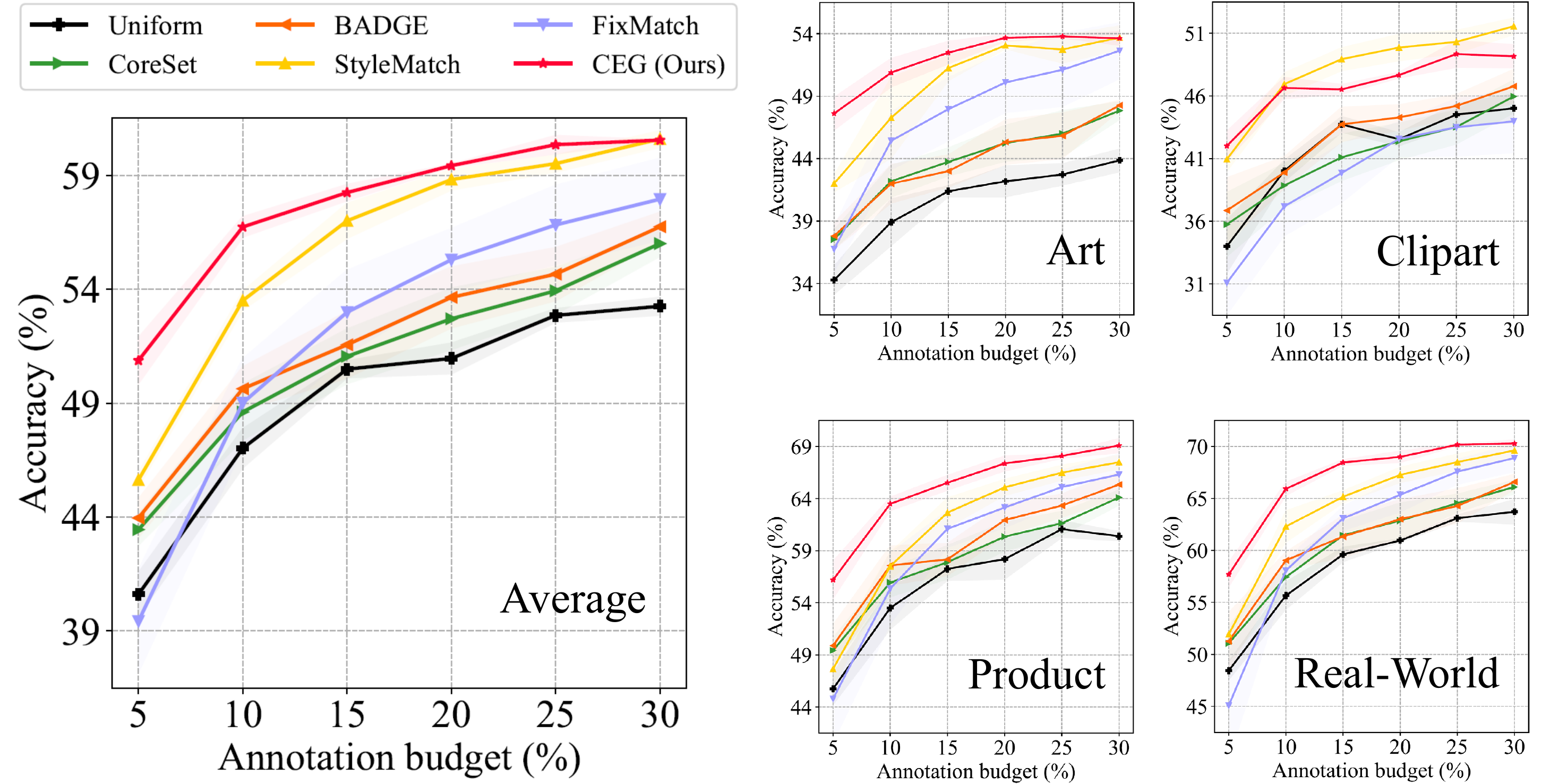}
    \caption{Comparisons with \textbf{increasing annotation budget}.}\label{fig:line-home}
\end{figure}

\begin{table*}[t]
\caption{\textbf{Ablation studies} on PACS and Office-Home datasets (\textbf{5\%} annotation). The best results are emphasized in bold.}
\label{table:ablation}
\centering
\resizebox{1\linewidth}{!}{
\begin{tabular}{cl|cccc|c||cccc|c}
\toprule
\multirow{2}{*}{Strategies} & \multirow{2}{*}{Cases} & \multicolumn{5}{c||}{PACS dataset (\%)} & \multicolumn{5}{c}{Office-Home dataset (\%)} \\
\cline{3-12}
& & Art & Cartoon & Photo & Sketch & Average & Art & Clipart & Product & Real-World & Average \\
\midrule
\multirow{4}{*}{\shortstack{Active\\Exploration}}
& w/ Uniform & 76.42\scriptsize{$\pm$1.27} & 69.92\scriptsize{$\pm$3.01} & 87.09\scriptsize{$\pm$2.03} & 71.92\scriptsize{$\pm$4.97} & 76.34\scriptsize{$\pm$2.05} 
& 45.25\scriptsize{$\pm$1.29} & 40.48\scriptsize{$\pm$2.18} & 53.48\scriptsize{$\pm$2.00} & 55.95\scriptsize{$\pm$1.08} & 48.79\scriptsize{$\pm$0.96} \\
& w/o $S_{u}$  & 80.00\scriptsize{$\pm$2.08} & 67.56\scriptsize{$\pm$2.34} & 89.43\scriptsize{$\pm$0.71} & 71.11\scriptsize{$\pm$5.64} & 77.03\scriptsize{$\pm$1.28} 
& 46.12\scriptsize{$\pm$2.04} & 40.84\scriptsize{$\pm$0.98} & 52.56\scriptsize{$\pm$0.75} & 55.83\scriptsize{$\pm$1.47} & 48.84\scriptsize{$\pm$0.84} \\
& w/o $S_{r}$ & 76.77\scriptsize{$\pm$2.74} & 67.91\scriptsize{$\pm$4.38} & 90.90\scriptsize{$\pm$1.19} & 72.46\scriptsize{$\pm$1.44} & 77.01\scriptsize{$\pm$0.74} 
& 46.14\scriptsize{$\pm$1.05} & 39.59\scriptsize{$\pm$1.25} & 55.99\scriptsize{$\pm$0.82} & 57.02\scriptsize{$\pm$2.06} & 49.68\scriptsize{$\pm$0.71} \\
& w/o $S_{d}$ & 78.21\scriptsize{$\pm$0.97} & 68.95\scriptsize{$\pm$2.54} & 91.03\scriptsize{$\pm$1.50} & 72.58\scriptsize{$\pm$3.24} & 77.69\scriptsize{$\pm$0.61} 
& 45.78\scriptsize{$\pm$2.34} & 40.79\scriptsize{$\pm$1.78} & 54.39\scriptsize{$\pm$1.46} & \textbf{58.24}\scriptsize{$\pm$1.83} & 49.84\scriptsize{$\pm$0.82} \\
\midrule
\multirow{6}{*}{\shortstack{Semi-Supervised\\Generalization}}
& w/o $\mathcal{L}_{ac}$ w/o $\mathcal{L}_{eg}$ & 62.61\scriptsize{$\pm$2.81} & 68.51\scriptsize{$\pm$2.34} & 82.84\scriptsize{$\pm$3.85} & 48.65\scriptsize{$\pm$4.32} & 65.65\scriptsize{$\pm$1.35} 
& 38.51\scriptsize{$\pm$1.06} & 33.62\scriptsize{$\pm$1.50} & 48.60\scriptsize{$\pm$2.59} & 49.89\scriptsize{$\pm$5.07} & 42.66\scriptsize{$\pm$1.44} \\
& w/o $\mathcal{L}_{ac}$ & 76.16\scriptsize{$\pm$2.11} & 67.05\scriptsize{$\pm$3.37} & 88.07\scriptsize{$\pm$1.80} & 60.37\scriptsize{$\pm$5.36} & 72.91\scriptsize{$\pm$1.31} 
& 44.65\scriptsize{$\pm$1.69} & 37.99\scriptsize{$\pm$1.25} & 55.41\scriptsize{$\pm$2.16} & 56.42\scriptsize{$\pm$2.09} & 48.62\scriptsize{$\pm$0.59} \\
& w/o $\mathcal{L}_{eg}$ & 72.13\scriptsize{$\pm$1.43} & 68.09\scriptsize{$\pm$3.37} & 86.03\scriptsize{$\pm$4.02} & \textbf{75.12}\scriptsize{$\pm$2.31} & 75.34\scriptsize{$\pm$0.84} 
& 40.38\scriptsize{$\pm$2.15} & 34.68\scriptsize{$\pm$1.29} & 46.10\scriptsize{$\pm$2.20} & 47.94\scriptsize{$\pm$1.66} & 42.27\scriptsize{$\pm$1.18} \\
& w/o $\mathcal{L}_{eg}$ ($\mathcal{M}^{intra}$) & 76.52\scriptsize{$\pm$2.01} & 67.69\scriptsize{$\pm$3.50} & 89.89\scriptsize{$\pm$1.72} & 74.72\scriptsize{$\pm$2.88} & 77.20\scriptsize{$\pm$1.53} 
& 45.85\scriptsize{$\pm$2.72} & 38.50\scriptsize{$\pm$1.47} & 54.12\scriptsize{$\pm$2.65} & 55.80\scriptsize{$\pm$1.66} & 48.57\scriptsize{$\pm$1.36} \\
& w/o $\mathcal{L}_{eg}$ ($\mathcal{M}^{inter}$) & 75.36\scriptsize{$\pm$2.59} & 68.20\scriptsize{$\pm$2.66} & 91.18\scriptsize{$\pm$1.05} & 72.49\scriptsize{$\pm$2.28} & 76.81\scriptsize{$\pm$1.32} 
& 47.47\scriptsize{$\pm$3.43} & 38.97\scriptsize{$\pm$2.26} & 52.99\scriptsize{$\pm$2.77} & 55.10\scriptsize{$\pm$1.48} & 48.63\scriptsize{$\pm$1.32} \\
& w/ static $T$ & 76.86\scriptsize{$\pm$2.37} & 69.61\scriptsize{$\pm$1.63} & 90.79\scriptsize{$\pm$1.13} & 73.96\scriptsize{$\pm$2.06} & 77.81\scriptsize{$\pm$1.09} 
& 45.90\scriptsize{$\pm$1.76} & 40.46\scriptsize{$\pm$2.33} & 55.90\scriptsize{$\pm$1.50} & 56.21\scriptsize{$\pm$2.30} & 49.62\scriptsize{$\pm$1.02} \\
\midrule
\multicolumn{2}{c|}{\textbf{CEG}} & \textbf{80.12}\scriptsize{$\pm$0.37} & \textbf{71.11}\scriptsize{$\pm$0.96} & \textbf{92.32}\scriptsize{$\pm$1.68} & 73.13\scriptsize{$\pm$2.87} & \textbf{79.17}\scriptsize{$\pm$0.83} & \textbf{47.60}\scriptsize{$\pm$1.32} & \textbf{42.01}\scriptsize{$\pm$1.19} & \textbf{56.20}\scriptsize{$\pm$1.79} & 57.69\scriptsize{$\pm$1.18} & \textbf{50.87}\scriptsize{$\pm$0.99}\\
\bottomrule
\end{tabular}}
\end{table*}

\begin{figure*}[t]
    \centering
    \includegraphics[trim={0cm 0cm 0cm 0cm},clip,width=1.76\columnwidth]{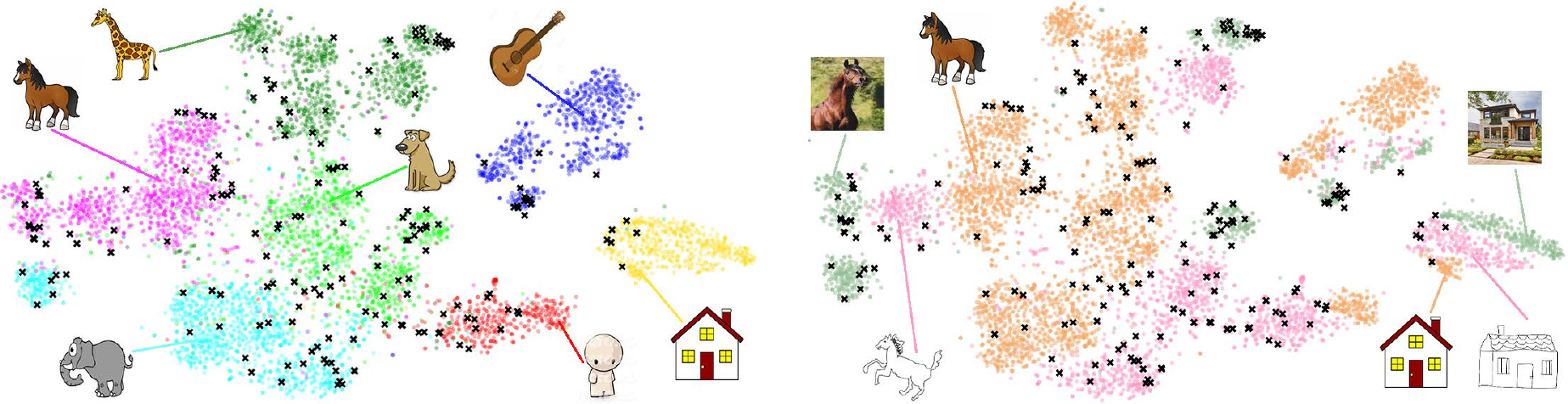}
    \caption{\textbf{T-SNE visualization \cite{Maaten2008VisualizingDU} of the selected samples}. Different colors represent different classes (\textbf{left}) and domains (\textbf{right}). Each dot is the embedding of a sample and the \textbf{black crosses are the selected samples}. 
    Dataset: PACS; Target domain: Art.}\label{fig:tsne}
\end{figure*}

\begin{figure}[t]
    \centering
    \includegraphics[trim={0cm 0cm 0cm 0cm},clip,width=0.99\columnwidth]{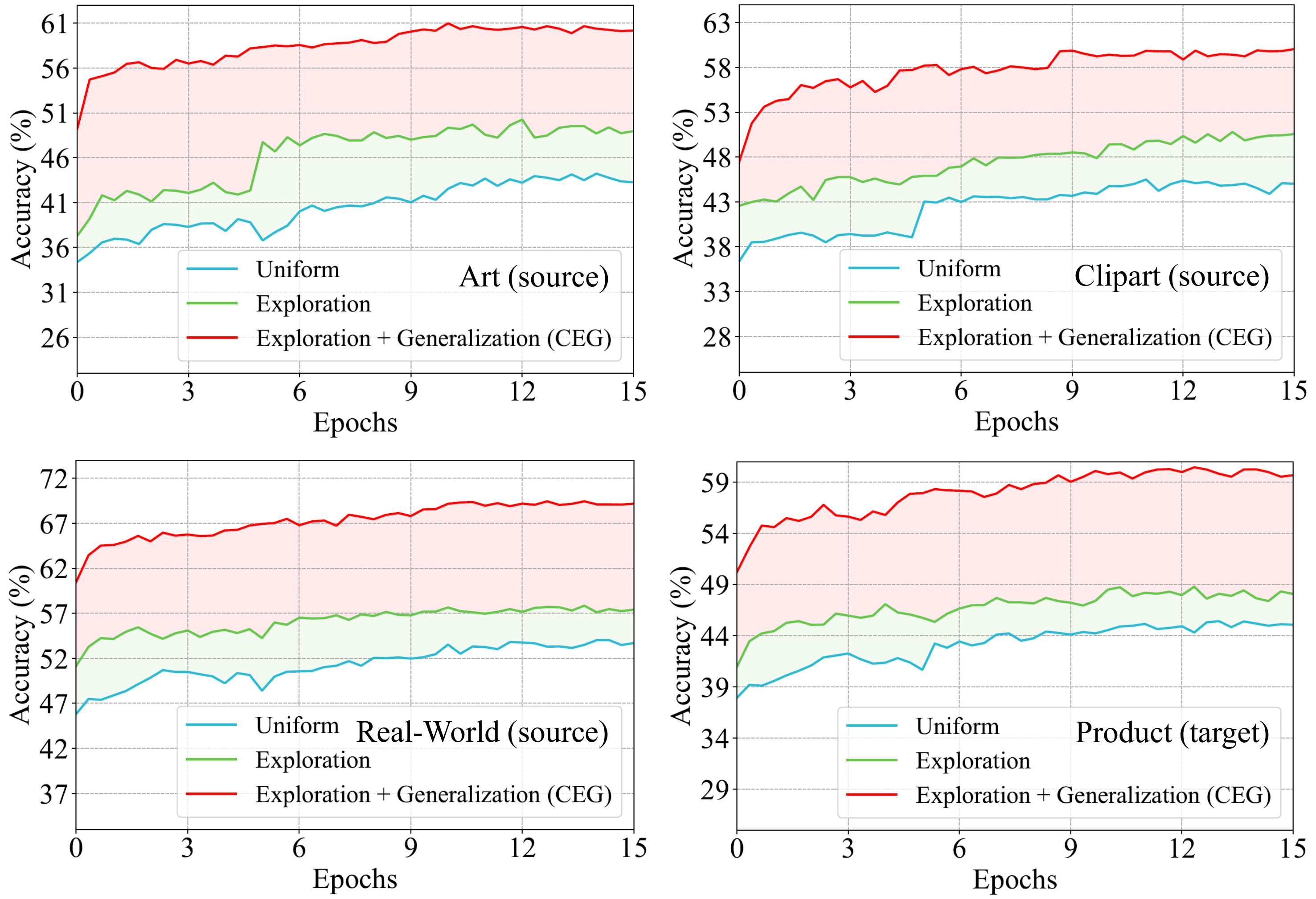}
    \caption{\textbf{Accuracy} of w/ uniform, w/ active exploration, w/ active exploration + semi-supervised training (CEG).}\label{fig:acc}
\end{figure}

In this section, we first evaluate our framework \textbf{CEG} in label-limited scenarios, and then give sensitivity analysis of hyper-parameters, ablation studies of the components, and in-depth empirical analysis. 

\textbf{Datasets.}
We adopt two popular pubilc datasets that are \textbf{PACS} \cite{li2017deeper} and \textbf{Office-Home} \cite{venkateswara2017deep}. PACS contains 7 categories within 4 domains, i.e., \textit{Art}, \textit{Cartoon}, \textit{Sketch}, and \textit{Photo}. Office-Home has 65 classes in 4 domains, i.e., \textit{Art}, \textit{Clipart}, \textit{Product}, and \textit{Real-World}. 

\textbf{Baseline methods.}
We implement four types of baselines.
(1) Domain generalization (\textbf{DG}): DeepAll (training with mixed multi-source data), JiGen \cite{Carlucci2019DomainGB}, CrossGrad \cite{shankar2018generalizing}, DDAIG \cite{zhou2020deep}, DAEL \cite{zhou2021domain3}, RSC \cite{HuangWXH20}, and FACT \cite{xu2021fourier}.
(2) Active learning (\textbf{AL}): Uniform (uniform selection), Entropy \cite{wang2014new}, BvSB \cite{joshi2009multi}, Confidence \cite{wang2014new}, CoreSet \cite{sener2017active}, and BADGE \cite{ash2019deep}.
(3) Semi-supervised learning (\textbf{SSL}): MeanTeacher \cite{tarvainen2017mean}, MixMatch \cite{berthelot2019mixmatch}, and FixMatch \cite{Sohn2020FixMatchSS}.
(4) Semi-supervised domain generalization (\textbf{SSDG}): StyleMatch \cite{zhou2021semi}. 
See Section \ref{sec:rel} for details.

\textbf{Implementation details.}
Following \cite{Carlucci2019DomainGB, HuangWXH20, xu2021fourier}, we use a pre-trained ResNet-18 \cite{he2016deep} as the backbone and conduct leave-one-domain-out experiments by choosing one domain to hold out as the target domain. For fair comparisons, we implement all the methods with the same settings, i.e., SGD optimizer with learning rate 0.003 for feature extractor and 0.01 for classifier, pre-training/learning epochs on PACS and Office-Home datasets are 30/30 and 15/15, respectively, and batch-size is 16, et al. 
In experiments, we directly use ``$T$'' to represent ``$T^{fin}$'' in Equation (\ref{equ:T}) for simplicity. We adopt the percentage of the unlabeled samples for $T$ instead of a distance value. We set $T^{ini}=\frac{T}{2}$. 
The hyper-parameters $\{\delta,T,\gamma_{1},\gamma_{2}\}$ are set to $\{0.3,50\%,3,1\}$ and $\{0.1, 70\%, 0.5, 0.5\}$ for PACS and Office-Home, respectively. 
Half annotation budget is used as the initial budget to initialize the labeled dataset. 
We report the results over five runs. 

\subsection{Main Results of CEG}
\textbf{CEG vs DG methods.} 
Table \ref{table:dg-pacs} and \ref{table:dg-home} report the results with 5\% annotation budget on PACS and Office-Home datasets, respectively. 
We observe that the accuracy of DG methods drops rapidly when only 5\% labeled data is given. In comparison, our method CEG can select the informative data to label and utilize both the labeled and unlabeled data to boost generalization performance in this challenging label-limited scenario. Most notably, CEG can even achieve competitive results with only 5\% annotation budget compared to the DG methods with full annotation on the PACS dataset. It reveals that CEG generally realizes label-efficient domain generalization by exploiting only a small quota of labeled data and massive unlabeled data. We attribute this success to the effective collaboration mechanism between active exploration and semi-supervised generalization, which unleashes the latent power of the limited annotation budget. 
Since the DG methods may not be good at tackling the label-limited task as they can only use the labeled data, we further compare our CEG method with AL, SSL, and SSDG methods.  

\textbf{CEG vs AL, SSL, SSDG methods.} 
Table \ref{table:result} reports the results with 5\% annotation budget on PACS and Office-Home datasets. CEG outperforms other methods on half of the tasks and yields the best average accuracy on the PACS dataset. It is probably because the AL and SSL methods rely on the i.i.d. assumption, and the SSDG method does not label and exploit the important source data. In contrast, CEG selects the most informative samples for query via active exploration, and hence captures multi-source distribution and boosts generalization ability more accurately. Besides, the performance of CEG is significantly better than other methods on the Office-Home dataset. We attribute it to the construction of domain-class knowledge centroids, which greatly helps CEG to precisely explore unknown regions during active exploration, and effectively expand knowledge and generalize domain invariance during semi-supervised generalization on the Office-Home dataset (because Office-Home has 65 classes but PACS only has 7 classes).

\subsection{Sensitivity Analysis}
As shown in Figure \ref{fig:sen}, CEG is generally robust to the hyper-parameters and outperforms other methods even with the default settings, i.e., $\delta=1.0$ (78.79\% on PACS and 46.34\% on Office-Home), $T=100\%$ (78.09\% on PACS and 50.49\% on Office-Home), $\gamma_{1}=\gamma_{2}=1.0$ (78.12\% on PACS and 50.37\% on Office-Home), indicating that exhaustive hyper-parameter fine-tuning is not necessary for CEG to achieve excellent performance in label-efficient generalization learning. 

\subsection{Results with Increasing Annotation Budget}
Figure \ref{fig:line-home} shows the results with increasing annotation on Office-Home dataset. CEG consistently outperforms other methods by sharp margins on the average accuracy and three of the four tasks. The significant performance achieved by CEG when given a low budget is probably due to the query based-active exploration, but this advantage could be weakened when given a higher budget. 

\subsection{Why does CEG Work?}
\textbf{Ablation studies} are reported in Table \ref{table:ablation}. 
The three criteria of active exploration, i.e., uncertainty $S_{u}$, representativeness $S_{r}$, and diversity $S_{d}$, are all important for learning multi-source distribution, and the integration of them further make full use of the limited annotation, compared with uniform selection. 
For semi-supervised generalization, both knowledge expansion and generalization $\mathcal{L}_{eg}$ and augmentation consistency $\mathcal{L}_{ac}$ are necessary to yield remarkable results. The intra- and inter-domain knowledge augmentation datasets, i.e., $\mathcal{M}^{intra}$ and $\mathcal{M}^{inter}$, both play vital roles in improving generalization performance. It is noteworthy that the proposed $\mathcal{L}_{eg}$ significantly improves average accuracy from 42.27\% to 50.87\% on Office-Home. Besides, the devised dynamic threshold $T$ shows its effectiveness of learning with increasing difficulty compared to the static one.
The above results illustrate that each component is indispensable, and the exploration and generalization complement and promote each other for achieving the excellent performance. 

T-SNE visualization is shown in Figure \ref{fig:tsne}. The left figure shows class-ambiguous samples, i.e., the samples distribute on class boundary, are selected for learning class discriminability. The right figure shows that, in general, the selected samples distribute uniformly and representatively in each domain, illustrating the effectiveness of the domain representativeness and information diversity criteria. These three criteria help CEG to select the most informative samples for learning multi-source distribution, which facilitates the generalizable model training in semi-supervised generalization.

Accuracy curve on Office-Home dataset is shown in Figure \ref{fig:acc}. 
Active exploration selects the most important samples and grasps the key knowledge of multi-source distribution to effectively improve performance on each domain, compared with uniform sample selection. Semi-supervised generalization further markedly boosts performance by expanding the obtained knowledge and generalizing domain invariance. They promote each other to achieve remarkable generalization performance on the target domain.

\section{Conclusion}
\label{sec:con}
We introduce a practical task named label-efficient domain generalization, and propose a novel method called CEG for this task via active exploration and semi-supervised generalization. The two modules promote each other to improve model generalization with the limited annotation. 
In future work, we may extend our method to a more challenging setting that domain labels are unknown. 



\begin{acks}
This work was supported in part by National Key Research and Development Program of China (2021YFC3340300), Young Elite Scientists Sponsorship Program by CAST (2021QNRC001), National Natural Science Foundation of China (No. 62006207, No. 62037001), Project by Shanghai AI Laboratory (P22KS00111), the Starry Night Science Fund of Zhejiang University Shanghai Institute for Advanced Study (SN-ZJU-SIAS-0010), Natural Science Foundation of Zhejiang Province (LZ22F020012), and the Fundamental Research Funds for the Central Universities (226-2022-00142).
\end{acks}

\bibliographystyle{ACM-Reference-Format}
\bibliography{sample-base}

\end{document}